\documentclass[runningheads]{llncs}
\usepackage{graphicx}

\usepackage{tikz}
\usepackage{comment}
\usepackage{amsmath,amssymb} 
\usepackage{color}
\usepackage{multirow}
\usepackage{overpic}
\usepackage{bbding}
\usepackage{enumitem}
\usepackage{booktabs}
\usepackage[breaklinks=true,bookmarks=false,colorlinks,citecolor=black,linkcolor=black]{hyperref}

\def\eg{{\em e.g.}}
\def\ie{{\em i.e.}}
\def\etal{{\em et al.}}
\newcommand{\minitab}[2][l]{\begin{tabular}{#1}#2\end{tabular}}

\begin{document}
\pagestyle{headings}
\mainmatter
\def\ECCVSubNumber{5256}  

\title{Unbiased Multi-Modality Guidance \\for Image Inpainting} 

\titlerunning{Unbiased Multi-Modality Guidance for Image Inpainting}
%
\author{Yongsheng Yu\inst{1,3} \and
Dawei Du\inst{2} \and
Libo Zhang\inst{1,3,4}\thanks{Corresponding author (libo@iscas.ac.cn).} \and
Tiejian Luo\inst{3}
}
\authorrunning{Y. Yu et al.}
%
\institute{Institute of Software Chinese Academy of Sciences, China \and
Kitware, USA \and
University of Chinese Academy of Sciences, China \and
Nanjing Institute of Software Technology, China \\
\email{yuyongsheng19@mails.ucas.ac.cn}; \email{cvdaviddo@gmail.com}; \email{libo@iscas.ac.cn}; \email{tjluo@ucas.ac.cn}
}
\maketitle

\begin{abstract}
Image inpainting is an ill-posed problem to recover missing or damaged image content based on incomplete images with masks. Previous works usually predict the auxiliary structures (\eg, edges, segmentation and contours) to help fill visually realistic patches in a multi-stage fashion. However, imprecise auxiliary priors may yield biased inpainted results. Besides, it is time-consuming for some methods to be implemented by multiple stages of complex neural networks. To solve this issue, we develop an end-to-end multi-modality guided transformer network, including one inpainting branch and two auxiliary branches for semantic segmentation and edge textures. Within each transformer block, the proposed multi-scale spatial-aware attention module can learn the multi-modal structural features efficiently via auxiliary denormalization. Different from previous methods relying on direct guidance from biased priors, our method enriches semantically consistent context in an image based on discriminative interplay information from multiple modalities. Comprehensive experiments on several challenging image inpainting datasets show that our method achieves state-of-the-art performance to deal with various regular/irregular masks efficiently.
\keywords{biased prior, multi-modality guidance, auxiliary denormalization, image inpainting}
\end{abstract}

\section{Introduction}
\label{sec:intro}
Image inpainting aims to repair missing or damaged image content based on known information of an image. It has been applied on many real-world scenarios, such as image editing~\cite{DBLP:conf/icpr/ArdinoL0LN20,DBLP:journals/tog/BarnesSFG09}, unwanted object removal~\cite{DBLP:conf/cvpr/CriminisiPT03,DBLP:conf/nips/ShettyFS18}, and old photo restoration~\cite{DBLP:conf/aaai/SongCSHH19}.

Following the assumption that corrupted images have adequate knowledge for inpainting~\cite{DBLP:conf/cvpr/Yu0YSLH18,DBLP:conf/cvpr/LiWZDT20}, modern image inpainting methods~\cite{DBLP:conf/cvpr/PathakKDDE16,DBLP:conf/iccvw/NazeriNJQE19,DBLP:conf/cvpr/XiongYLYLBL19,DBLP:conf/eccv/LiaoXWLS20,DBLP:conf/cvpr/LiWZDT20} employ an encoder-decoder architecture. Concretely, they focus on various contextual attention mechanisms to learn the known visible content and fill the missing region. However, this assumption does not hold if the image is damaged by larger masks. It is difficult to provide sufficient semantically consistent information for realistic image inpainting based on known area in a RGB image. 

\begin{figure*}[t]
  \centering
  \includegraphics[width=0.9\textwidth]{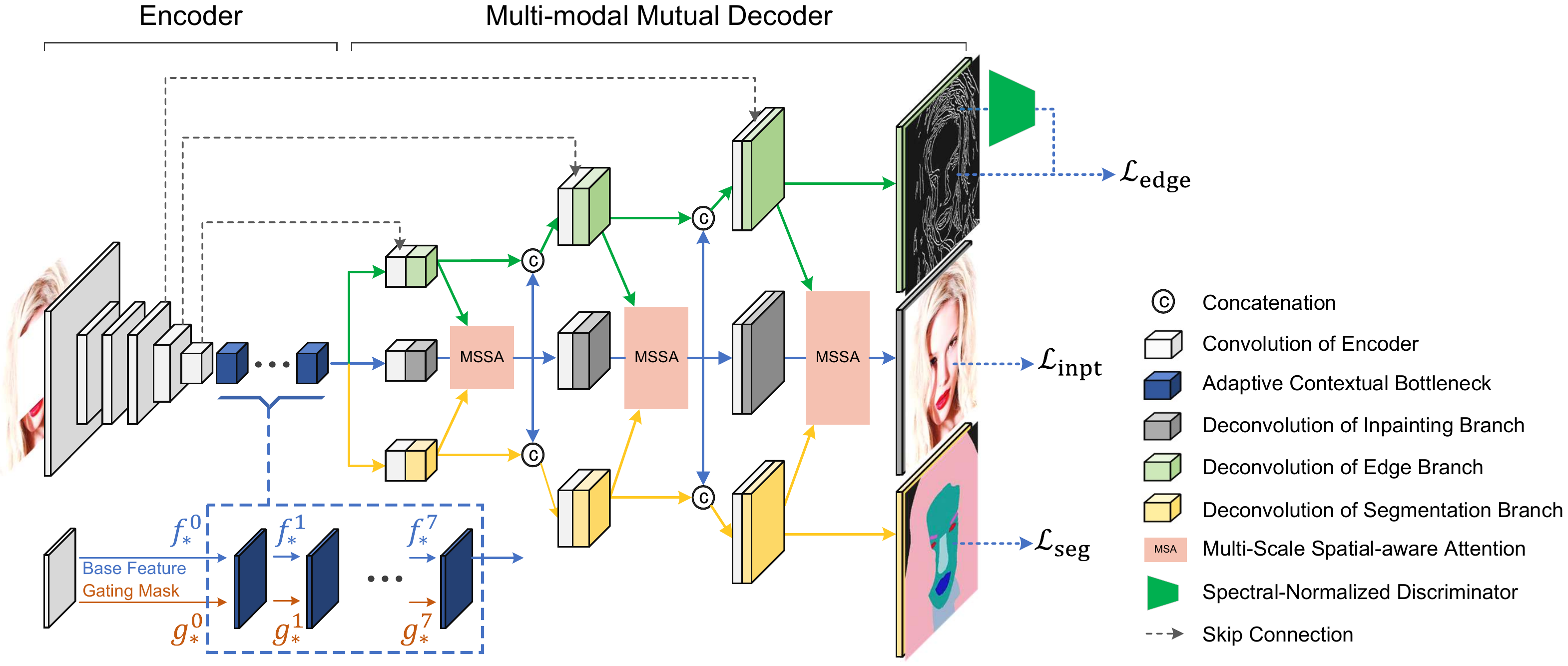}
  \caption{Architecture of our Multi-Modality guided Transformer that couples various modalities including RGB image, semantic segmentation, and edge textures.}\label{fig:framework}
\end{figure*}

Therefore, recent approaches~\cite{DBLP:conf/cvpr/XiongYLYLBL19,DBLP:conf/iccvw/NazeriNJQE19,DBLP:conf/bmvc/SongYSWHK18,DBLP:conf/cvpr/Liao00L021,DBLP:journals/corr/abs-2103-15087} have made great efforts to introduce auxiliary priors, such as \textit{edges}, \textit{segmentation}, and \textit{contours}, to facilitate improving image inpainting performance. However, they still suffer from the biased prior issue by using predicted auxiliary structures to guide image inpainting intermediately. Without ground-truth in testing phase, such direct guidance is inevitably biased, resulting in more deviations and errors for image inpainting. On the other hand, previous works~\cite{DBLP:conf/bmvc/SongYSWHK18,DBLP:conf/iccv/LiuJX019} are usually divided into multiple stages of neural networks under the U-Net architecture. If each stage contains a complex subnetwork, it is time-consuming for potential real-world inpainting applications. This problem becomes more prominent when extending to video inpainting. For example, Liu~\etal~\cite{DBLP:conf/iccv/LiuJX019} tackle the image inpainting problem by a two-stage process, \ie, two individual U-Nets for rough inpainting and refinement inpainting, yielding the running speed of only $1.37$ FPS.
  
To solve the above issues, we propose a new multi-modality guided transformer network for image inpainting. As shown in Fig. \ref{fig:framework}, it follows the U-Net style~\cite{DBLP:conf/miccai/RonnebergerFB15} encoder-decoder architecture. 
In the encoder, we first develop the adaptive contextual bottlenecks for better context reasoning. To adapt to the current image content and missing region, the gating mask is updated to weight different dilated convolutions to enhance base features. 
Then, the multi-modal mutual decoder is proposed to decode the enhanced features into three modalities, \ie, RGB image, and corresponding semantic segmentation and edge textures. It consists of one image inpainting branch and two auxiliary branches for semantic segmentation and edge textures. Unlike existing approaches based on direct guidance from predicted auxiliary structures, we focus on jointly learning the unbiased discriminative interplay information among the three branches. 
Specifically, the proposed multi-scale spatial-aware attention mechanism integrates multi-modal feature maps via auxiliary denormalization to reduce duplicated and noisy content for image inpainting. Supervised by ground-truth RGB images, semantic segmentation and edge maps, the whole network is trained in an end-to-end fashion efficiently. Note that segmentation and edge annotations can be provided by the off-the-shelf algorithms~\cite{DBLP:conf/eccv/ChenZPSA18,DBLP:conf/iccvw/NazeriNJQE19}. 

\begin{figure*}[t]
   \centering
   \begin{overpic}[width=0.9\textwidth]{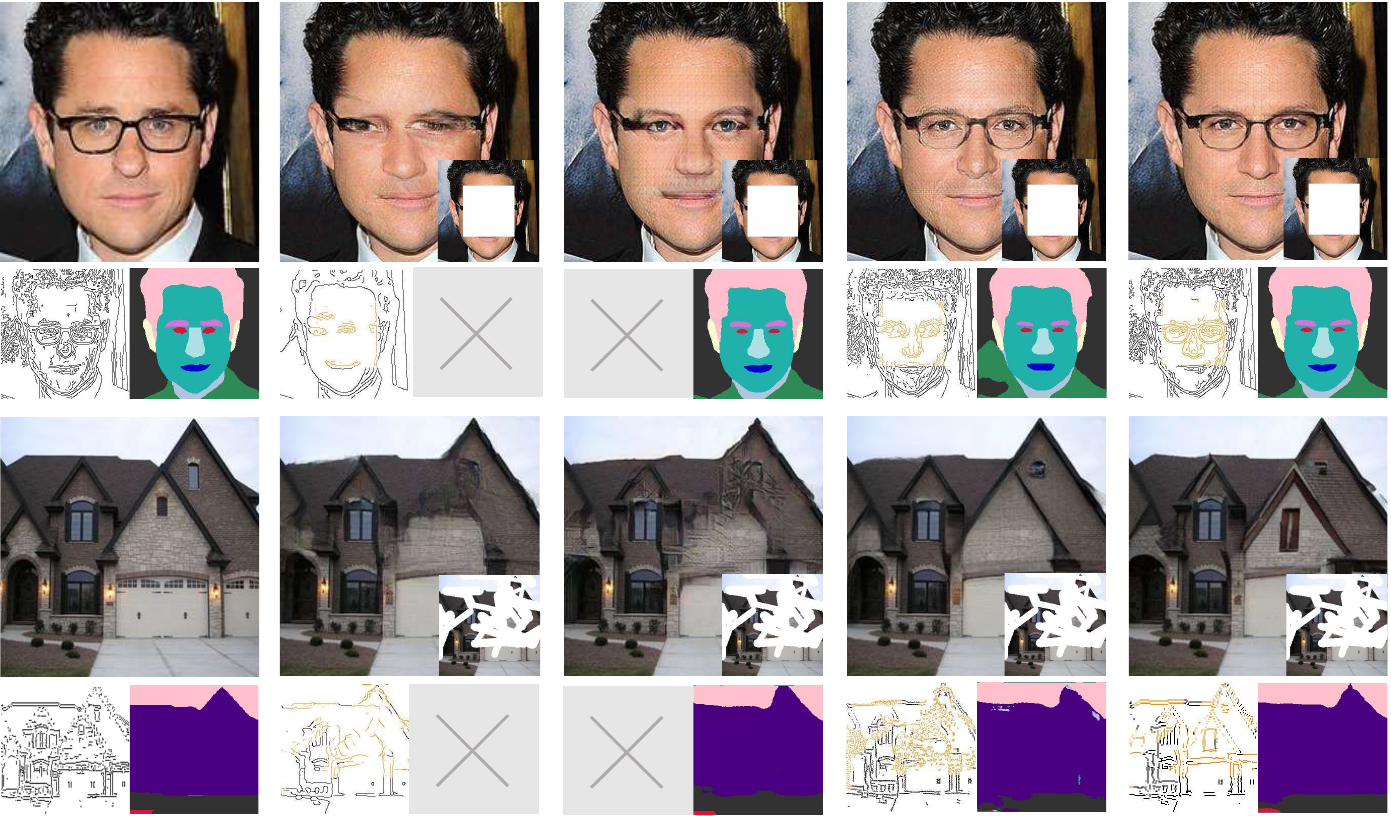} \small
   \put(6,-2.5){GT}
   \put(25,-2.5){EC~\cite{DBLP:conf/iccvw/NazeriNJQE19}}
   \put(45,-2.5){SPG~\cite{DBLP:conf/bmvc/SongYSWHK18}}
   \put(68,-2.5){Ours$^\ast$}
   \put(89,-2.5){Ours}
   \end{overpic}
   \caption{Influence of biased prior guidance. \XSolid~means no edge prior for SPG~\cite{DBLP:conf/bmvc/SongYSWHK18} and segmentation prior for EC~\cite{DBLP:conf/iccvw/NazeriNJQE19}. Ours$^\ast$ denotes the variant of our multi-modality guided image inpainting method with inaccurate edge and segmentation priors by reducing the loss weights of two auxiliary branches by $30$ times.}\label{fig:motivation}
\end{figure*}
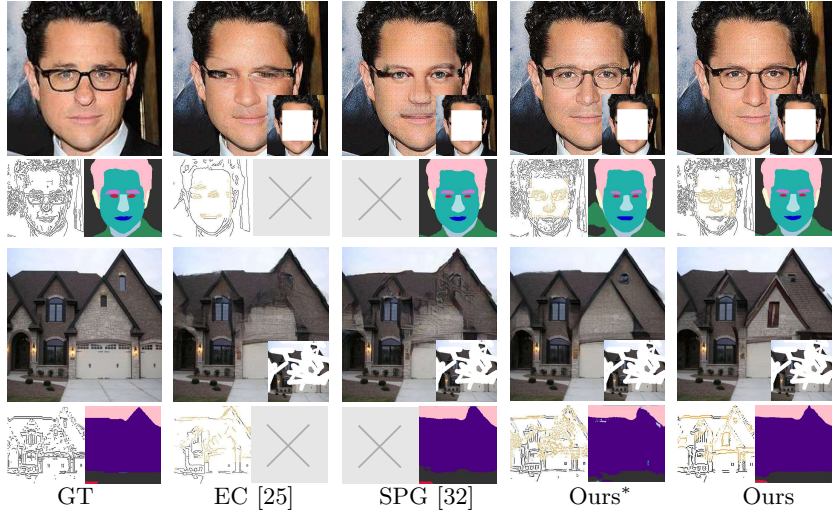

As shown in Fig. \ref{fig:motivation}, previous image inpainting methods fail to restore correct faces and buildings based on either biased edge~\cite{DBLP:conf/iccvw/NazeriNJQE19} or segmentation~\cite{DBLP:conf/bmvc/SongYSWHK18} prior. On the contrary, our method still achieves robust results even though the glasses are not repaired in edge prior (see Ours$^\ast$ in the 1st row of Fig. \ref{fig:motivation}) or the roof shape is not predicted correctly in segmentation prior (see Ours$^\ast$ in the 2nd row of Fig. \ref{fig:motivation}). It demonstrates that our method can extract discriminative unbiased context information to guide image inpainting.
To verify the effectiveness of our method, the experiment is conducted on three datasets including CelebA-HQ~\cite{DBLP:conf/iclr/KarrasALL18,DBLP:conf/cvpr/Lee0W020}, OST~\cite{DBLP:conf/cvpr/WangYDL18} and CityScapes~\cite{DBLP:conf/cvpr/CordtsORREBFRS16}. The results show our method achieves the state-of-the-art image inpainting performance. For example, our method obtains the best FID score on the CelebA-HQ dataset with both regular and irregular masks, yeilding $\sim2$ gain over the second best performer CTSDG~\cite{guo2021image}. By using segmentation results from DeepLabv3+~\cite{DBLP:conf/eccv/ChenZPSA18}, our method still performs well on those datasets without segmentation annotation (\eg, Places2 \cite{DBLP:journals/pami/ZhouLKO018}).

\textbf{Contributions.} 1. We propose an end-to-end multi-modality guided transformer to learn interplay information from multiple modalities including RGB image, edge textures and semantic segmentation. 2. We develop the multi-scale spatial-aware attention mechanism with auxiliary denormalization to capture compact and discriminative multi-modal features to guide unbiased image inpainting. 3. Comprehensive results on several benchmarks demonstrate the effectiveness of our unbiased multi-modality guidance, especially for irregular masks.

\section{Related Work}
{\flushleft {\bf Image inpainting.}}
Mainstream image inpainting methods employ the encoder-decoder architecture based on the U-Net~\cite{DBLP:conf/miccai/RonnebergerFB15}. For example, Pathak~\etal~\cite{DBLP:conf/cvpr/PathakKDDE16} introduces an adversarial network~\cite{DBLP:journals/corr/GoodfellowPMXWOCB14} to help train the U-Net and mitigate the blurring caused by the pixel-level averaging property of a reconstruction loss. After that, Contextual Attention (CA)~\cite{DBLP:conf/cvpr/Yu0YSLH18} is a two-stage coarse-to-fine model to weight known region as the reference of mission region. Using partial conv~\cite{DBLP:conf/eccv/LiuRSWTC18}, Recurrent Feature Reasoning (RFR)~\cite{DBLP:conf/cvpr/LiWZDT20} applies multiple iterations at the bottleneck of the encoder from outside to inside for large corrupt areas. Different from partial conv~\cite{DBLP:conf/eccv/LiuRSWTC18} with a heuristic mask update step to standard convolution, Gated Conv (GC)~\cite{DBLP:conf/iccv/YuLYSLH19} improves this mask update process with a learnable convolution layer.

To better exploit context between missing and uncorrupted regions, Iizuka~\etal~\cite{DBLP:journals/tog/IizukaS017} first introduces multiple residual modules~\cite{DBLP:conf/cvpr/HeZRS16} of dilation convolution~\cite{DBLP:conf/cvpr/YuKF17} as the bottleneck in the encoder. However, it may bring the ``gridding'' problem~\cite{DBLP:journals/corr/ChenPSA17,DBLP:conf/wacv/WangCYLHHC18} due to only sampling padded non-zero positions. That is, a single constant dilation rate results in either sparse convolution kernels (large hole rate) or difficulty crossing over large masks (small hole rate). To this end, Wang \etal~\cite{DBLP:conf/nips/WangTQSJ18} develop a generative multi-column network for image inpainting. Recently, Zeng~\etal~\cite{DBLP:journals/corr/abs-2104-01431} propose the AOT blocks to aggregate contextual transformations from varying receptive fields, which capture both informative distant image contexts and rich patterns of interest. Different from above methods, we introduce a new adaptive contextual bottleneck in the encoder, where the dynamic gating updating weights different pathways of dilated convolutions based on various masks. 

{\flushleft {\bf Image inpainting with auxiliary structures.}}
Due to the ill-posed nature of reconstructing missing regions, additional structural priors (\eg, \textit{edges}, \textit{segmentation}, and \textit{contours}) are used to facilitate image inpainting models for more realistic results. Edge Connect (EC)~\cite{DBLP:conf/iccvw/NazeriNJQE19} relies on the corrupted canny edge image to deliver finer inpainting results. Cao and Fu~\cite{DBLP:journals/corr/abs-2103-15087} introduce an extra encoder to infer precise wireframe sketches to bypass the pool coherence of canny edge. According to the style and spatial consistency of semantic segmentation, Segmentation Prediction and Guidance network (SPG)~\cite{DBLP:conf/bmvc/SongYSWHK18} is a two-stage based segmentation and RGB image inpainting model, where DeepLabv3+~\cite{DBLP:conf/eccv/ChenZPSA18} is used to estimate the segmentation of corrupted image. Another work~\cite{DBLP:conf/cvpr/XiongYLYLBL19} is a new three-stage based model to locate and fill foreground object and its contour by disentangling the inter-object intersection.

However, the above multi-stage methods are usually time-consuming. For better efficiency, the Semantic Guidance and Evaluation (SGE) network~\cite{DBLP:conf/eccv/LiaoXWLS20} couples with segmentation and image inpainting at different layers of decoder, where the segmentation after completing and confidence scoring guides image inpainting by semantic normalization~\cite{DBLP:conf/cvpr/Park0WZ19}. Liao~\etal~\cite{DBLP:conf/cvpr/Liao00L021} propose the Semantic-wise Attention Propagation (SWAP) module to capture the semantic relevance between segmentation and image textures in non-local operation. Recently, Yang~\etal~\cite{DBLP:conf/aaai/YangQS20} predict explicit edge embedding with an attention mechanism to facilitate image inpainting by the multi-task learning strategy. It worth mentioning that most aforementioned works use estimated auxiliary structures as the direct guidance of image inpainting. On the contrary, we develop the multi-head spatial-aware attention module to guide image inpainting based on jointly learned discriminative features from unbiased auxiliary priors.

{\flushleft {\bf Transformers in image inpainting.}}
Inspired by Vision Transformer \cite{DBLP:conf/iclr/DosovitskiyB0WZ21}, recent methods~\cite{DBLP:conf/mm/DengHZMW21,DBLP:conf/mm/YuZWPCLMXM21} decode the long-range dependencies between input features for better image inpainting. Deng \etal~\cite{DBLP:conf/mm/DengHZMW21} learn relations between the corrupted and uncorrupted regions and exploit their respective internal closeness. Yu~\etal~\cite{DBLP:conf/mm/YuZWPCLMXM21} introduce the bidirectional autoregressive transformer that enables bidirectionally modeling of contextual information of missing regions. In contrast, our method propose a new multi-modality guided transformer to capture interplay information across three modalities.

\section{Multi-Modality Guided Transformer}
The original image $\mathbf{I}$ is degraded as a corrupted image $\mathbf{I}_m=\mathbf{I} \odot (1-\mathbf{M})$, where the pixel values in the missing region $\mathbf{M}$ equal to $0$ are defined as invisible pixels. Our goal is to produce semantically reasonable and visually realistic reconstructed images $\mathbf{I}_\text{pred}$ with the input of the corrupted image $\mathbf{I}_m$. Similar to previous works~\cite{DBLP:conf/cvpr/PathakKDDE16,DBLP:journals/tog/IizukaS017,DBLP:conf/iccv/YuLYSLH19,DBLP:conf/cvpr/LiWZDT20}, we retain the U-Net style encoder-decoder architecture. As illustrated in Fig. \ref{fig:framework}, the multi-modality guided transformer contains an encoder with adaptive contextual bottlenecks, and a multi-modal mutual decoder with multi-scale spatial-aware attention, described in detail as follows.

\subsection{Encoder with Adaptive Contextual Bottlenecks}
\begin{figure}[t]
  \centering
  \includegraphics[width=0.9\columnwidth]{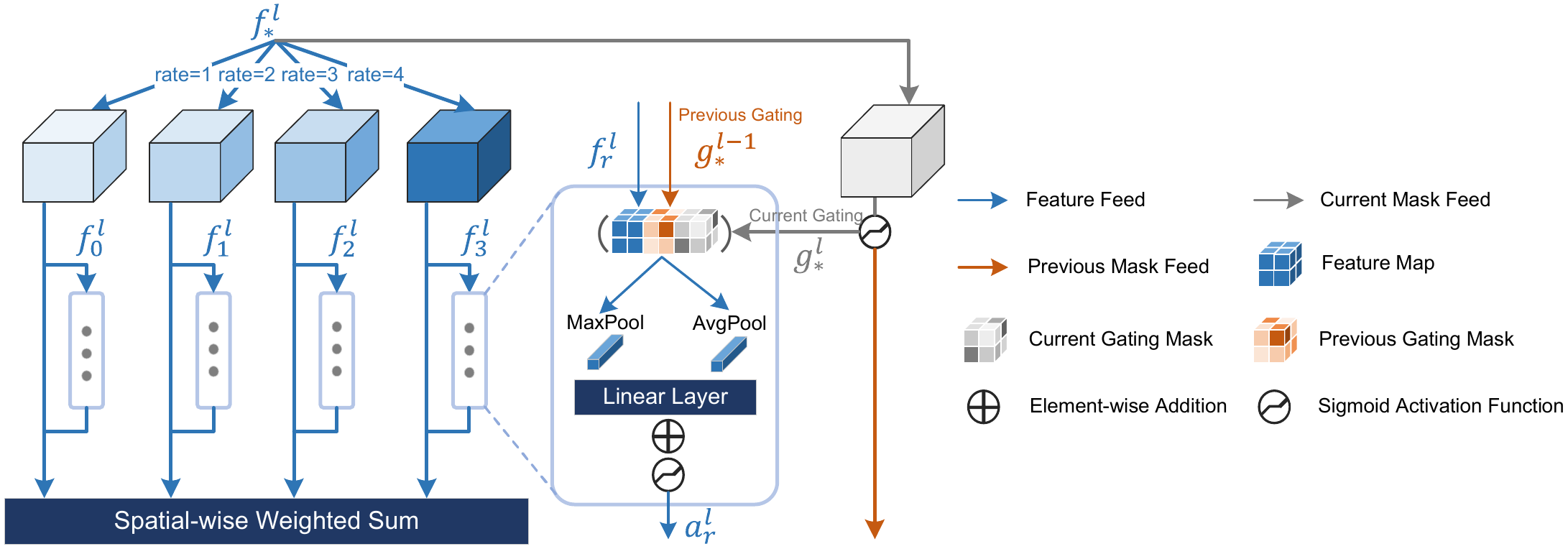}
  \caption{Structure of Adaptive Contextual Bottlenecks in the encoder.}\label{fig:aed}
\end{figure}

For better context reasoning, the multi-stream structure is used in the encoder to weight dilated convolutions and encode the current image content and missing region. Unlike simply stacking parameters in previous ASPP \cite{DBLP:journals/corr/ChenPSA17} and AOT \cite{DBLP:journals/corr/abs-2104-01431}, we develop a stack of Adaptive Contextual Bottlenecks (ACB) to adapt to the specific mask shape size and image context by dynamic gating. As shown in Fig. \ref{fig:aed}, the ACB module consists of four parallel pathways of convolutional layers with different dilation rate and one gating mask to weight dilated convolutions. In this way, the encoder can enlarge the perceptual field of convolutions and find the most plausible pathway according to the current missing region.

Given the corrupt image $\mathbf{I}_m$, the base features $f_{*}^{0}$ and gating $g_{*}^{0}$ are initialized by the last layer (gated conv) of encoder. Then $f_{*}^{l}$ and $g_{*}^{l}$ at each layer is updated by the ACB block.
The gating mask $g^l_\ast$ is used to estimate the probability of missing region based on the feature map at the $l$-th layer ($l=1,\cdots,L$), \ie, $g^l_\ast=\mathrm{gconv}(f^l_\ast)$, where $\mathrm{gconv}$ denotes the gated conv operation~\cite{DBLP:conf/iccv/YuLYSLH19}. In terms of each pathway with dilation rate $r$, we compute the dilated feature maps $f^l_r$ based on $f^l_\ast$ and corresponding weight $a^l_r$. Similar to \cite{DBLP:conf/eccv/WooPLK18}, the spatial-wise weight $a^l_r$ is calculated based on both average and max pooling of concatenation of dilated feature maps $f^l_r$ and gating masks $g^l_\ast,g^{l-1}_\ast$, \ie, 
$a^l_r = \sigma(\mathrm{fc}(\mathrm{avg}(g^l_r))+\mathrm{fc}(\mathrm{max}(g^l_r)))$, where $\sigma$ is the sigmoid function, and $\mathrm{avg}$ and $\mathrm{max}$ are the average and maximal pooling respectively. $\mathrm{fc}$ denotes the fully-connected layer, and the gating mask for each pathway is calculated as $g^l_r = \mathrm{conv}([f^l_r;g^l_\ast;g^{l-1}_\ast])$.
Finally, the feature map at the $(l+1)$-th ACB layer is updated by the spatial-wise weighted summation of $f^l_r$ as
\begin{equation}
f^{l+1}_\ast = \sum_{r \in R} \frac{\exp(a^l_r)}{\sum_{r\in R}\exp(a^l_r)} \cdot f^l_r + f^l_\ast,
\end{equation}
where $R$ denotes the set of different dilation rates. The fractional term denotes element-wise product between dilated feature map $f_{r}^{l}$ and attention vector $a_{r}^{l}$, weighting dilation block based on mask and image context. For simplicity, we omit the subscript $l$ in the following sections.

\subsection{Multi-modal Mutual Decoder}
Given enhanced features $f_\ast$, the decoder use stacks of transformer blocks to learn the structural multi-modal information jointly. It consists of three branches, \ie, one \textit{inpainting branch} to recover the damaged image, and two \textit{auxiliary branches} with additional segmentation and edge priors.

As shown in Fig. \ref{fig:framework}, within each transformer block, we first calculate the attention among feature maps from three branches by the proposed Multi-Scale Spatial-aware Attention (MSSA). Then, the enhanced features are split to combine the previous feature maps in each branch for attention calculation at next stage. Note that the skip connections between the encoder and decoder are used to prevent network degradation. After three stages, we predict the inpainted image $\mathbf{I}_\text{pred}$, edge and segmentation maps. Thus we leverage the structural features from auxiliary branches to enforce the model focus on discriminative interplay features for more realistic image inpainting. 

To learn mutual features from different modalities, it is intuitive to simply concatenate or add the feature maps in three branches. Nevertheless, such strategies may introduce duplicated and noisy content for image inpainting. To effectively integrate compact features from \textit{auxiliary branches}, we introduce a new Multi-Scale Spatial-aware Attention (MSSA) mechanism as follows. 
\begin{figure*}[t]
  \centering
  \includegraphics[width=0.9\textwidth]{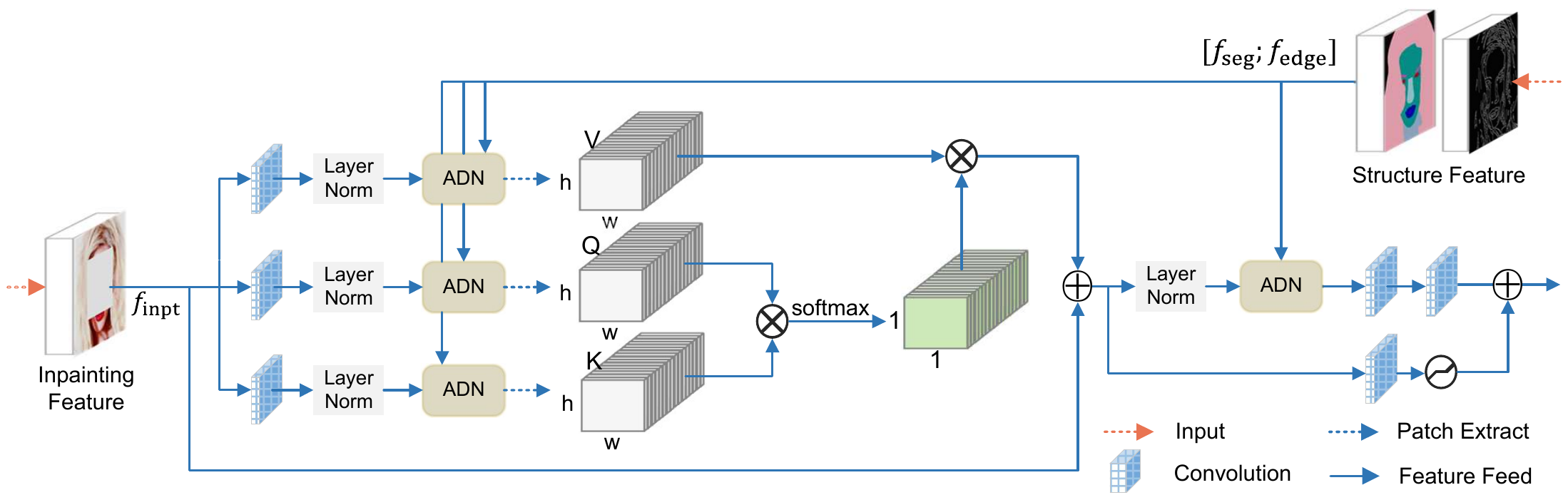}
  \caption{Illustration of Multi-Scale Spatial-aware Attention.}\label{fig:msa}
\end{figure*}

{\noindent {\bf Multi-scale spatial-aware attention.}}
Based on the encoded feature maps $f_\ast$, we use $f_\text{inpt}, f_\text{edge}, f_\text{seg}$ to denote the input feature maps for the inpainting branch, edge branch, and segmentation branch, respectively. As illustrated in Fig. \ref{fig:msa}, we combine the feature maps from three branches by the following Auxiliary DeNormalization (ADN): 
\begin{equation}
\mathrm{ADN}(f_\text{inpt} | [f_\text{edge}; f_\text{seg}]) = \gamma \odot \mathrm{LN}(f_\text{inpt}) + \beta,
\end{equation}
where $[;]$ denotes the matrix concatenation along channel dimension, and $\odot$ the element-wise multiplication. $\mathrm{LN}$ denotes layer normalization~\cite{DBLP:journals/corr/BaKH16}. $\gamma$ and $\beta$ are the affine transformation parameters learned by two convolutional layers based on $[f_\text{edge}; f_\text{seg}]$ (see the top-right corner of Fig. \ref{fig:msa}). In this way, the multi-modal features are merged based on context from auxiliary structures that varies with respect to different spatial location. 

Then, the merged features are embedded into query $Q$, key $K$ and value $V$. Similar to \cite{DBLP:conf/eccv/ZengFC20}, the embedded feature map is spatially split into $N$ patches, \ie, $P_i\in\mathbb{R}^{h\times w\times c} (i=1,\dots,N)$, where $h,w,c$ denote the height, width and channel of patches respectively. The normalized self-attention $\alpha_{i,j}$ between patches $i$ and $j$ can be calculated as $\alpha_{i,j}= \mathrm{softmax}(\frac{Q_i \cdot K_j^T}{\sqrt{h\cdot w\cdot c}}), \quad i, j \in {1, \dots, N}$. 
Note that we can perform multi-head self-attention like~\cite{DBLP:conf/iclr/DosovitskiyB0WZ21}. Thus the feature map of each patch is updated in a non-local form, \ie, $\hat{P}_i = \sum_{j=1}^{N} \alpha_{i,j} V_j$. 

\begin{figure*}[t]
	\centering
	\includegraphics[width=0.85\textwidth]{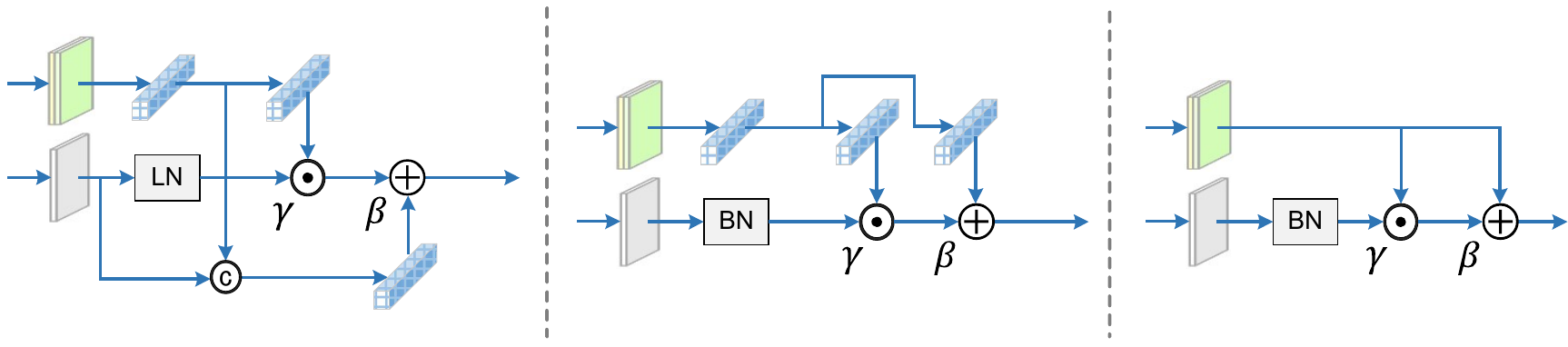}
	\caption{(a) Our Auxiliary DeNormalization (ADN). (b) SPatially-Adaptive DEnormalization (SPADE)~\cite{DBLP:conf/cvpr/Park0WZ19}. (c) Adaptive Instance Normalization (AaIN)~\cite{DBLP:conf/iccv/HuangB17}. LN, BN and IN denote layer, batch and instance normalizations respectively.}
	\label{fig:adn}
\end{figure*}

{\noindent {\bf Comparison between existing denormalization methods.}}
Our ADN is related to two previous denormalization methods including AdaIN~\cite{DBLP:conf/iccv/HuangB17} and SPADE~\cite{DBLP:conf/cvpr/Park0WZ19}. As shown in Fig. \ref{fig:adn}, we compare the networks of three denormalization methods. However, they are different in two aspects:
\begin{itemize}
\item AdaIN~\cite{DBLP:conf/iccv/HuangB17} and SPADE~\cite{DBLP:conf/cvpr/Park0WZ19} learn the affine transformation parameters $\{\gamma,\beta\}$ based on the predicted auxiliary structures. Without ground-truth in testing phase, the predicted auxiliary structures are inevitably biased and result in inferior performance. In contrast, our ADN is based on the multi-modal features from two auxiliary branches.
\item AdaIN~\cite{DBLP:conf/iccv/HuangB17} leverages the image's mean and variance instead of learnable affine parameters. SPADE~\cite{DBLP:conf/cvpr/Park0WZ19} learns the spatial style of features by two convolutions after Batch Normalization. However, we combine features from both inpainting and auxiliary branches to learn the affine parameters.
\end{itemize}

{\noindent {\bf Gated feed-forward.}}
Finally, we piece all feature maps $\hat{P}_i$ together and reshape them with the original scale of input inpainting features $f_\text{inpt}$. Following the gated feed-forward layer, we can output the final feature maps for inpainted image prediction. Similar to gated conv~\cite{DBLP:conf/iccv/YuLYSLH19}, the gated feed-forward layer can ease the color discrepancy problem by detecting potentially corrupted and uncorrupted regions.

\subsection{Optimization}
To train our network, the overall loss consists of three terms, \ie,
\begin{equation}
    \mathcal{L} = \mathcal{L}_\text{inpt} + \lambda_\text{edge} \mathcal{L}_\text{edge} + \lambda_\text{seg} \mathcal{L}_\text{seg},
    \label{equ:all}
\end{equation}
where $\mathcal{L}_\text{inpt}$, $\mathcal{L}_\text{edge}$ and $\mathcal{L}_\text{seg}$ denote the loss terms for inpainting branch, edge branch and segmentation branch respectively. $\lambda_\text{edge}$ and $\lambda_\text{seg}$ are the balancing factors. The inpainting loss $\mathcal{L}_\text{inpt}$ follows the work in~\cite{DBLP:conf/eccv/LiuRSWTC18}. Similar to~\cite{DBLP:conf/iccvw/NazeriNJQE19}, we use both binary cross-entropy and adversarial loss functions to train the edge branch, \ie, 
\begin{equation}
\mathcal{L}_\text{edge} = w_1 \mathcal{L}_\text{BCE} + \mathcal{L}_\text{adv},
\label{equ:edge}
\end{equation}
where $w_1$ is the balancing weight. $\mathcal{L}_\text{BCE} = \frac{1}{N}\sum_{i=1}^N -[\mathbf{C}^i_\text{gt} \log \mathbf{C}^i_\text{pred} + (1-\mathbf{C}^i_\text{gt}) \log (1-\mathbf{C}^i_\text{pred})]$ predicts the edge structure, and $\mathcal{L}_\text{adv}=-\mathbb{E}\left[\mathbf{D}\left(\mathbf{C}_{\text{pred}}\right)\right]$ justifies if the predicted edge is fake or real. 
$\mathbf{C}_\text{pred}$ is the probability map between $0$ and $1$ for the reconstructed edge while $\mathbf{C}_\text{gt}$ is the ground-truth edge based on the canny operator~\cite{DBLP:conf/iccvw/NazeriNJQE19}. $\mathbf{D}$ denotes the spectral normalization discriminator \cite{DBLP:conf/iclr/MiyatoKKY18} that is composed of five convolutional layers.
For the segmentation branch, we use the cross-entropy loss denoted by $\mathcal{L}_\text{seg}=\frac{1}{N}\sum_{i=1}^N -\mathbf{S}^i_\text{gt} \log \mathbf{S}^i_\text{pred}$, where $\mathbf{S}^i_\text{gt}$ and $\mathbf{S}^i_\text{pred}$ denote the ground-truth category and predicted probability for pixel $i$.

\section{Experiment}
We compare our method with state-of-the-arts on three large-scale datasets. An extensive ablation study is conducted to investigate the important designs in our model. All experiments are conducted on two 24G TITAN RTX GPUs. 

{\flushleft {\bf Datasets.}}
CelebA-HQ dataset~\cite{DBLP:conf/iclr/KarrasALL18,DBLP:conf/cvpr/Lee0W020} is a large-scale face image dataset with $30K$ HD face images, where each image has a semantic segmentation mask corresponding to $19$ facial categories. 
Outdoor dataset (OST)~\cite{DBLP:conf/cvpr/WangYDL18} includes $9,900$ training images and $300$ testing images for $8$ semantic categories, which are obtained from the outdoor scene photography collection. 
Cityscapes dataset~\cite{DBLP:conf/cvpr/CordtsORREBFRS16} contains $5,000$ street view images belonging to $20$ categories. We expand the number of training images in this dataset, \ie, $2,975$ images from the training set and $1,525$ images from the test set are used for training, and $500$ images from the validation set are used for testing.
In addition, the Places2 dataset \cite{DBLP:journals/pami/ZhouLKO018} contains $10$ million images covering more than $400$ different types of scenes. 
We generate both regular and irregular masks to verify the ability of image inpainting methods. For regular masks, we draw a $128 \times 128$ centered square mask for CelebA-HQ and OST, and a $96 \times 96$ centered square mask for Cityscape. For irregular masks, we settle masks from~\cite{DBLP:conf/cvpr/LiWZDT20} for CelebA-HQ and masks from~\cite{DBLP:conf/eccv/LiuRSWTC18} for Cityscape and OST. 

\begin{table}[t]
\centering
\scriptsize
\caption{Quantitative comparison with the state-of-the-art approaches on CelebA-HQ. Easy, medium, and hard irregular masks denote the mask with coverage ratio of $10\%\sim20\%$, $30\%\sim40\%$, and $50\%\sim60\%$, respectively. $\uparrow$ higher is better, and $\downarrow$ lower is better. Best and second best results are \textbf{highlighted} and \underline{underlined}.}
\label{tab:main}
\setlength{\tabcolsep}{5mm}{
\begin{tabular}{c|c|ccc|c}
\hline
\multicolumn{2}{c|}{\multirow{2}{*}{mask type}}        & \multicolumn{3}{c|}{irregular}             & \multirow{2}{*}{regular} \\ \cline{3-5}
\multicolumn{2}{c|}{}                             & easy & medium & \multicolumn{1}{l|}{hard} &                          \\ \hline
\multicolumn{1}{c|}{\multirow{6}{*}{PSNR$\uparrow$}} & 
GC~\cite{DBLP:conf/iccv/YuLYSLH19}    & 29.30 & 25.72 & 23.77 & 25.75             \\
\multicolumn{1}{c|}{}                      &  RFR~\cite{DBLP:conf/cvpr/LiWZDT20}    & 29.22  & 26.12 & 24.31  &  24.85  \\
\multicolumn{1}{c|}{}                      &  CMGAN~\cite{DBLP:conf/iclr/ZhaoCSDLCX21}   & 29.06  & 25.79 &  23.90 & 24.33      \\
\multicolumn{1}{c|}{}                      &  ICT~\cite{Wan_2021_ICCV}    & 28.07 & 24.56  & 22.70 & 24.51    \\

\multicolumn{1}{c|}{}                      &  CTSDG~\cite{guo2021image} & \underline{29.59}  & \underline{26.59} & \underline{24.69} & \underline{26.56}  \\
\multicolumn{1}{c|}{}                      &  Ours &  \textbf{29.94}    &  \textbf{26.88}  &  \textbf{25.12}  &    \textbf{26.70}  \\ \hline
\multirow{6}{*}{SSIM$\uparrow$}                 &  GC~\cite{DBLP:conf/iccv/YuLYSLH19}    & 0.96 & 0.93 & 0.90 &   0.90  \\
                                          & RFR~\cite{DBLP:conf/cvpr/LiWZDT20}     & 0.96 & \underline{0.94} &  0.91 &   0.87  \\
                                          & CMGAN~\cite{DBLP:conf/iclr/ZhaoCSDLCX21} & \textbf{0.97} & \underline{0.94} & 0.91 & 0.87  \\
                                          & ICT~\cite{Wan_2021_ICCV}  & 0.96 & 0.92 & 0.89 & 0.87  \\
                                          & CTSDG~\cite{guo2021image} & \textbf{0.97} & \underline{0.94} & \underline{0.92}  & \underline{0.91}  \\
                                          & Ours &  \textbf{0.97}   &    \textbf{0.95}  & \textbf{0.93}  &   \textbf{0.92}                       \\ \hline
\multirow{6}{*}{FID$\downarrow$}              & GC~\cite{DBLP:conf/iccv/YuLYSLH19}    & 15.00 & 18.41 & 21.28 &  22.45 \\
                                          & RFR~\cite{DBLP:conf/cvpr/LiWZDT20} & 7.37 & 10.74  & 13.45 &  14.35 \\
                                          & CMGAN~\cite{DBLP:conf/iclr/ZhaoCSDLCX21} & 6.80 & 11.85 & 14.12 & 12.91 \\
                                          & ICT~\cite{Wan_2021_ICCV}  & \underline{6.54} & 11.80 & 15.93 & \underline{11.90} \\
                                          &  CTSDG~\cite{guo2021image} & 7.80 & \underline{10.14} & \underline{13.30}  & 14.52  \\
                                          & Ours &  \textbf{6.47}    &   \textbf{9.32}     &   \textbf{11.61}    &   \textbf{11.40}                       \\ \hline
\end{tabular}}
\end{table}

{\flushleft {\bf Evaluation Metrics.}}
Similar to the previous works \cite{DBLP:conf/eccv/LiaoXWLS20,DBLP:journals/corr/abs-2104-01431}, we use three metrics as follows. 
Peak Signal to Noise Ratio (PSNR) is an objective evaluation metric to assess the quality of generate images. 
Structural Similarity Index (SSIM)~\cite{DBLP:journals/tip/WangBSS04} uses the mean as an estimate of luminance, standard deviation as an estimate of contrast, and covariance as a measure of structural similarity to compare the difference between the generated and original images. 
Frechet Inception Distance (FID)~\cite{DBLP:conf/nips/HeuselRUNH17} evaluates the accuracy and diversity of generated images. Notably, the Inception network~\cite{DBLP:conf/nips/SalimansGZCRCC16} is used to extract the image features when calculating the FID score, and then calculate its mean and covariance matrix to estimate the distance between the ground-truth and generated data distribution. According to \cite{DBLP:conf/cvpr/ZhangIESW18}, deep metrics like FID are close to human perception.

\begin{figure*}[t]
   \centering
   \begin{overpic}[width=0.9\textwidth]{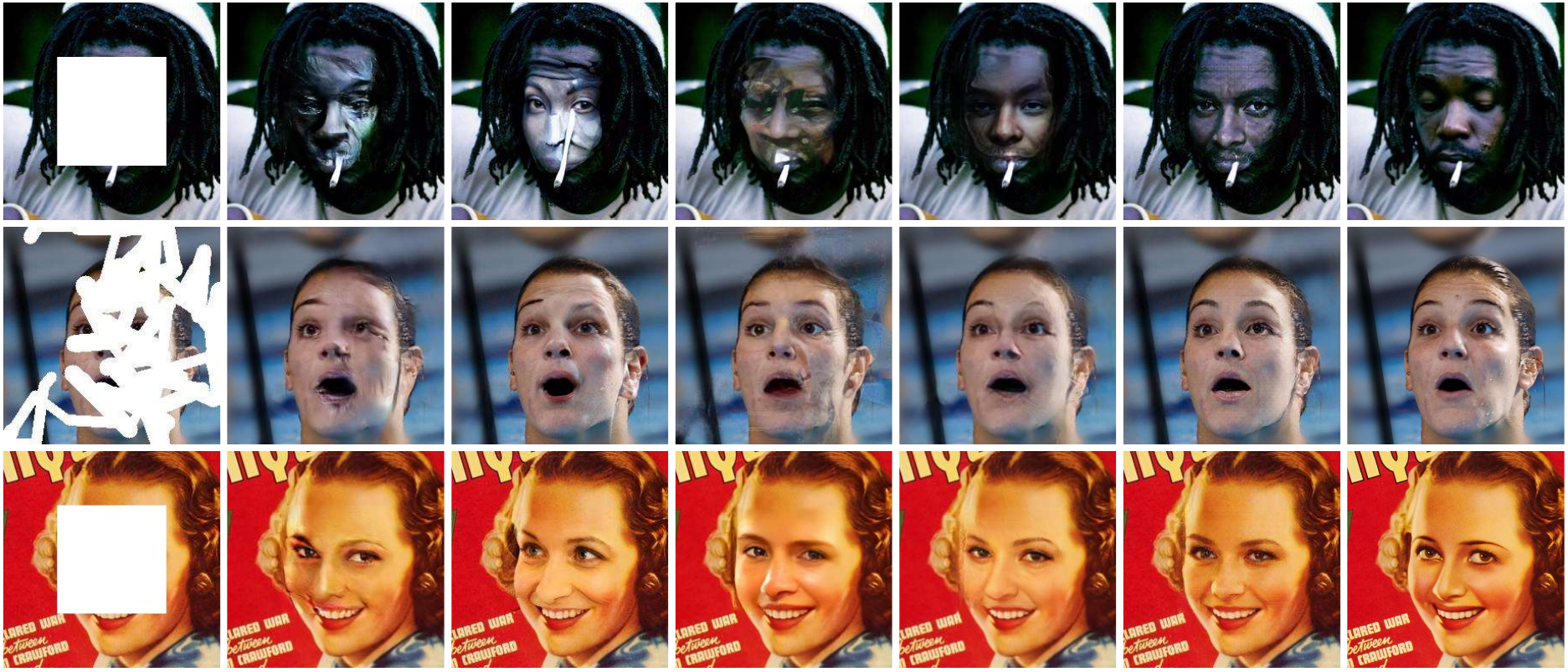} \small
   \put(2.5,-2){Masked}
   \put(19,-2){GC}
   \put(30,-2){CMGAN}
   \put(48,-2){ICT}
   \put(59.5,-2){CTSDG}
   \put(75.5,-2){Ours}
   \put(90.5,-2){GT}
   \end{overpic}
   \caption{Qualitative results of existing methods on CelebA-HQ.}\label{fig:VisualComparison}
\end{figure*}

\begin{figure*}[t]
   \centering
   \begin{overpic}[width=\textwidth]{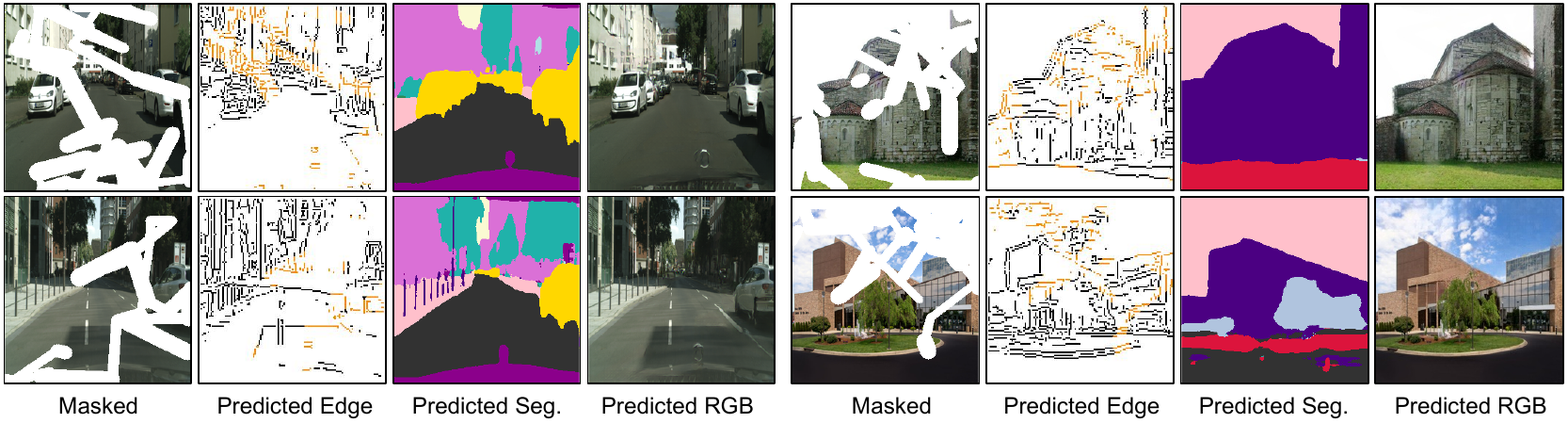} 
   \end{overpic}
   \caption{Qualitative results of our method on Cityscape (1 to 4 columns) and OST (5 to 8 columns).}\label{fig:VisualComparison2}
\end{figure*}

\subsection{Implementation Details}

Our model is supervised by auxiliary structures including edge textures and semantic segmentation. With regard to edge structure, we employ the canny detection method~\cite{DBLP:conf/iccvw/NazeriNJQE19} to generate edges of images. Besides, the CelebA-HQ, CityScapes and OST datasets all contain hand-crafted semantic segmentation, hence we can easily adopt these official labels for the segmentation part.
More details of implementation are shown in the supplementary.



\subsection{Result Analysis}
We compare our model with several state-of-the-art methods including GC~\cite{DBLP:conf/iccv/YuLYSLH19}, RFR~\cite{DBLP:conf/cvpr/LiWZDT20}, CMGAN~\cite{DBLP:conf/iclr/ZhaoCSDLCX21}, ICT~\cite{Wan_2021_ICCV}, CTSDG~\cite{guo2021image}, SPG~\cite{DBLP:conf/bmvc/SongYSWHK18}, SGINet~\cite{DBLP:conf/icpr/ArdinoL0LN20}, SGE~\cite{DBLP:conf/eccv/LiaoXWLS20}, and SWAP~\cite{DBLP:conf/cvpr/Liao00L021}. A quantitative comparison is carried out on three datasets in terms of both regular and irregular masks with different coverage ratios. Full comparison results~\cite{DBLP:journals/corr/abs-2104-01431,DBLP:conf/iccv/LiuJX019,DBLP:conf/cvpr/ZhengCC19,DBLP:conf/iccvw/NazeriNJQE19} we put in the appendix.

From Table~\ref{tab:main}, our method achieves the best or comparable performance among state-of-the-art image inpainting approaches that may not adopt auxiliary priors. Our method produces much better FID score than others for both regular and irregular masks, indicating that our inpainted results are more realistic.
In Table~\ref{tab:struct}, we compare several auxiliary prior guided inpainting approaches~\cite{DBLP:conf/iccvw/NazeriNJQE19,DBLP:conf/bmvc/SongYSWHK18,DBLP:conf/eccv/LiaoXWLS20,DBLP:conf/cvpr/Liao00L021}. For a fair comparison with the methods relying on only one auxiliary structure, we construct two variants, denoted by ``Ours w/o seg.'' and ``Ours w/o edge''. Compared with existing methods, our method achieves considerable gain respective to PSNR and FID especially on irregular masks. This is because our method focuses on the interplay representation from three modalities rather than directly guiding the image inpainting branch by predicted auxiliary structures (see Table \ref{tab:abla_att}).

In addition, we provide some visual examples on the CelebA-HQ dataset in Fig. \ref{fig:VisualComparison}. It can be seen that our method can generate more semantically consistent results compared with other approaches. More learned auxiliary priors of our method from CityScapes and OST datasets are visualized in Fig. \ref{fig:VisualComparison2}. 

\begin{table*}[t]
\centering
\small
\caption{Quantitative comparison with previous auxiliary prior guided approaches on OST and Cityscapes datasets.}
\label{tab:struct}
\resizebox{\textwidth}{15mm}{
\begin{tabular}{cccccccccccccc}
\hline
\multirow{3}*{\minitab[c]{method}}    &\multirow{3}*{\minitab[c]{auxiliary \\ prior}} & \multicolumn{6}{c}{OST}                                                                           & \multicolumn{6}{c}{CityScapes}                                                                    \\
    & & \multicolumn{3}{c}{regular}             & \multicolumn{3}{c}{irregular}             & \multicolumn{3}{c}{regular}             & \multicolumn{3}{c}{irregular}             \\ \cline{3-14} 
    & & PSNR$\uparrow$           & SSIM$\uparrow$          & FID$\downarrow$            & PSNR$\uparrow$           & SSIM$\uparrow$          & FID$\downarrow$            & PSNR$\uparrow$           & SSIM$\uparrow$          & FID$\downarrow$            & PSNR$\uparrow$           & SSIM$\uparrow$          & FID$\downarrow$            \\ \hline
EC~\cite{DBLP:conf/iccvw/NazeriNJQE19} &edge  & 19.32          & 0.76          & 41.25          & 19.12          & 0.74          & 42.27          & 21.71          & 0.76          & 19.87          & 17.63          & 0.72          & 39.04          \\
SPG~\cite{DBLP:conf/bmvc/SongYSWHK18} &seg. & 18.04          & 0.70          & 45.31          & 17.85          & 0.74          & 50.03          & 20.14          & 0.71          & 23.21          & 16.41          & 0.67          & 43.63          \\
SGINet~\cite{DBLP:conf/icpr/ArdinoL0LN20} &seg. & -          & -          & -          & -          & -          & -
 & 25.74 &0.87 &23.02 &18.53 &0.77 &57.53          \\
SGE~\cite{DBLP:conf/eccv/LiaoXWLS20} &seg.  
& 20.53          & \textbf{0.81}    &  40.67  
& 19.46          & 0.76          & 39.14
& 23.41          & 0.85    & 18.67 
& 17.78          & 0.74          & 41.45          \\
SWAP~\cite{DBLP:conf/cvpr/Liao00L021} &edge, seg. 
& 21.18    & \textbf{0.81} & \textbf{38.15} 
& 20.31 & 0.80    & 36.74 
& 23.89    & 0.84          & 18.14
& 17.86    & 0.76    & 38.18    \\
\hline
Ours w/o seg. &edge 
& 20.91 & 0.76 & 41.85
& 21.48 & 0.80 & 39.00
& 25.10 & 0.86 & 19.33
& 19.17 & \underline{0.78} & 37.50 \\
Ours w/o edge &seg. & 
\underline{21.80} & \underline{0.77} & 40.96 & 
\underline{22.58} & \underline{0.81} & \underline{36.03} & 
\underline{25.95} & \underline{0.87} & \underline{17.85} & 
\textbf{20.49} & \textbf{0.79} & \underline{34.79} \\
Ours &edge, seg. 
& \textbf{21.84} & \underline{0.77} & \underline{40.15}          
& \textbf{23.15} & \textbf{0.82} & \textbf{35.77}          
& \textbf{26.13} & \textbf{0.88} & \textbf{17.52}         
& \underline{20.43} & \textbf{0.79} & \textbf{33.45} \\ \hline
\end{tabular}
}
\end{table*}

\begin{figure}[t]
\centering
\begin{minipage}[t]{0.48\textwidth}
\centering
\includegraphics[width=\columnwidth]{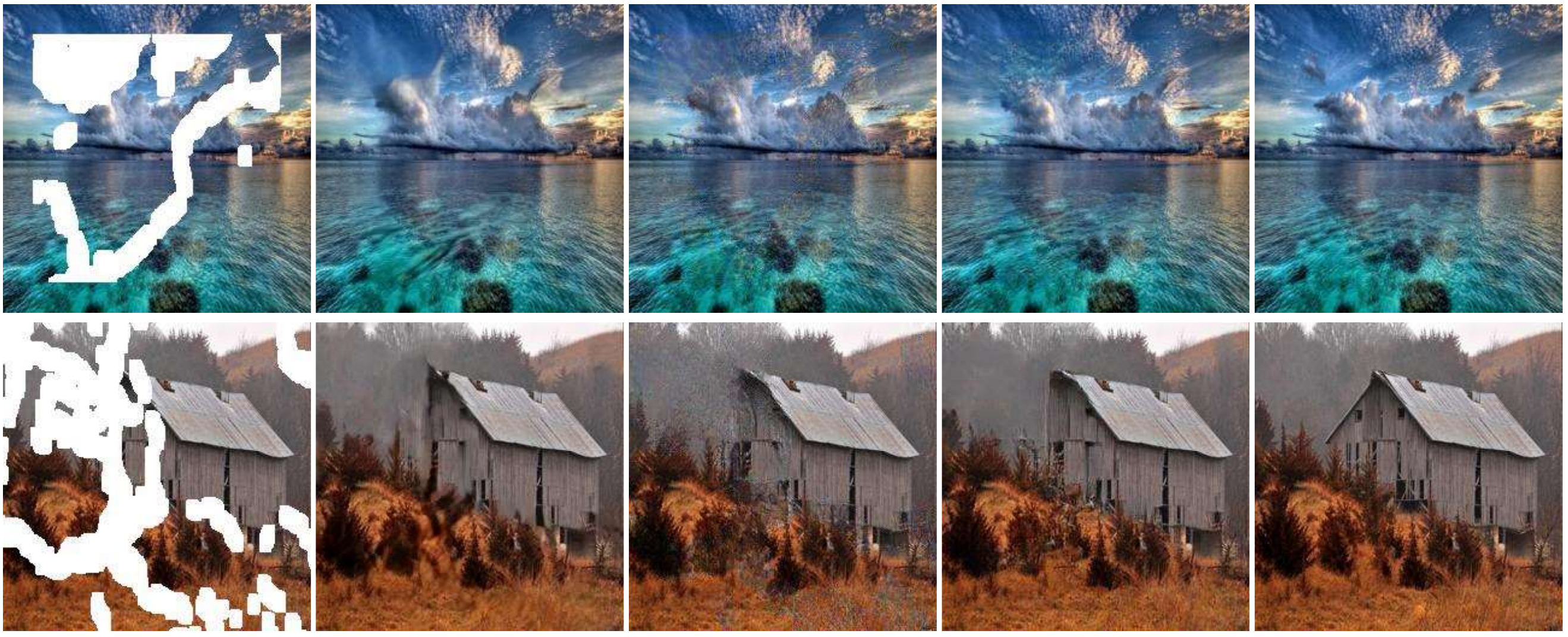}
\caption{Visual comparisons on Places2. From left to right: input, GC~\cite{DBLP:conf/iccv/YuLYSLH19}, EC~\cite{DBLP:conf/iccvw/NazeriNJQE19}, our method, and Ground Truth.}
\label{fig:PlacesComparison}
\end{minipage}
\begin{minipage}[t]{0.48\textwidth}
\centering
\includegraphics[width=\columnwidth]{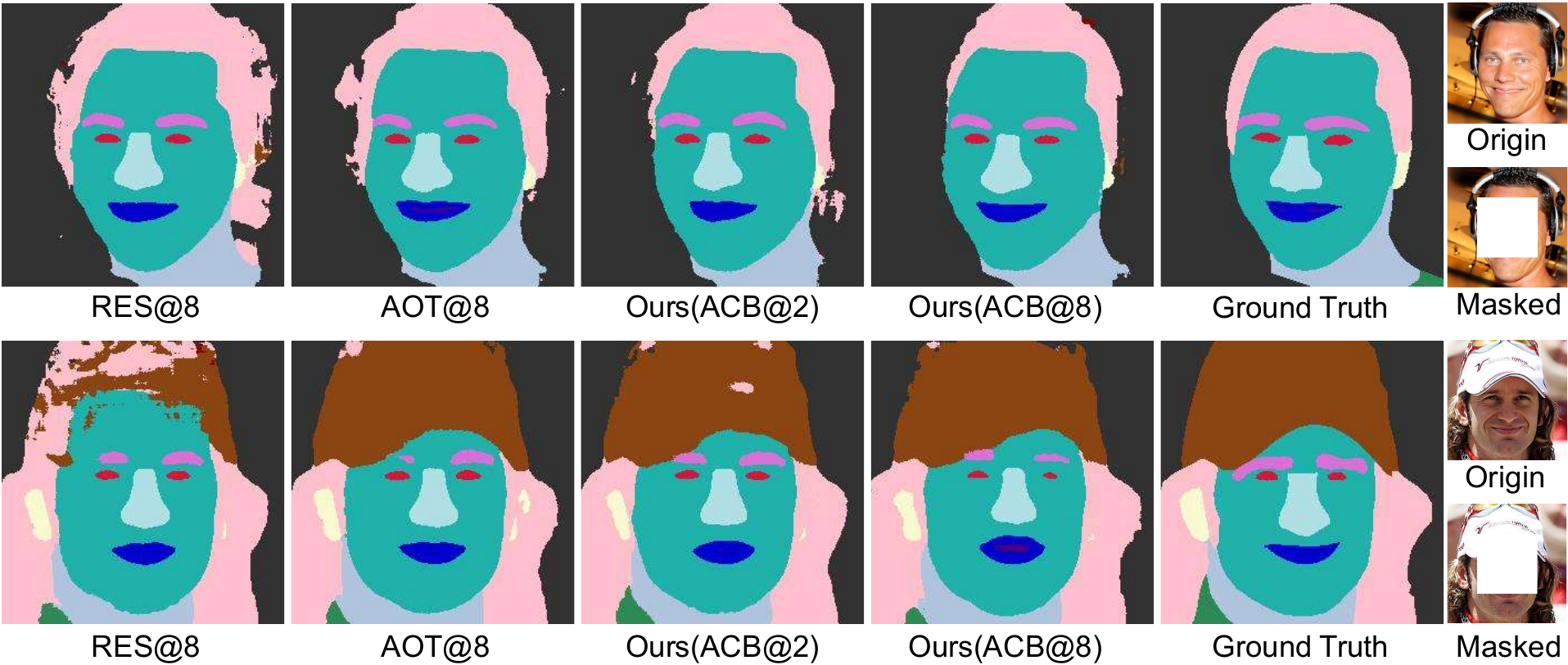}
\caption{Segmentation results with different bottlenecks on CelebA-HQ dataset with $128\times128$ regular center masks.}
\label{fig:vis_abla_aed}
\end{minipage}
\end{figure}

{\flushleft {\bf Additional results on Places2.}}
Similar to SGE \cite{DBLP:conf/eccv/LiaoXWLS20} and SWAP \cite{DBLP:conf/cvpr/Liao00L021}, we also conduct additional experiment on the Places2 dataset \cite{DBLP:journals/pami/ZhouLKO018} for a comprehensive evaluation. Since there is no ground-truth segmentation, we use the segmentation results by DeepLabv3+~\cite{DBLP:conf/eccv/ChenZPSA18} to supervise the segmentation branch in our model. As shown in Fig. \ref{fig:PlacesComparison}, the visual results show that our method still generate realistic inpainted images without ground-truth segmentation labels.

\subsection{Ablation Study}
To verify the effectiveness of the proposed modules in our network, the ablation experiments are carried out on the CelebA-HQ dataset.

\begin{table}[t]
    \caption{Contribution of two auxiliary branches in our method.}
    \label{tab:abla_struct}
    \footnotesize
    \centering
    \setlength{\tabcolsep}{10pt}
    \begin{tabular}{cc|ccc}
    \hline
    edge branch & segmentation branch & PSNR$\uparrow$ & SSIM$\uparrow$ & FID$\downarrow$ \\\hline
    \XSolidBrush   &  \XSolidBrush       &   25.88   &  0.90    & 12.36   \\ 
    \XSolidBrush  &    \Checkmark   &   26.47   &  0.91   & 11.42    \\
    \Checkmark   &   \XSolidBrush  &   26.19   &  0.90  & 11.95  \\
    \Checkmark   &   \Checkmark  &   \textbf{26.70}   &  \textbf{0.92}    & \textbf{11.40}    \\\hline
    \end{tabular}
\end{table}

{\flushleft {\bf Contribution of auxiliary branches.}}
In Table \ref{tab:abla_struct}, we construct three variants to verify the contribution of two auxiliary branches in our method. By learning from two auxiliary modalities, our method considerably outperforms the non-auxiliary variant w.r.t PSNR, SSIM, and FID. In addition, semantic segmentation contributes slightly more to image inpainting than edge textures. In summary, our Multi-Modal Mutual Decoder enriches semantic content on the inpainting branch by cross-attending segmentation and edge structures.

\begin{table}[t]
    \footnotesize
    \caption{Comparison with different attention mechanisms.}
    \label{tab:abla_att}
    \centering
    \setlength{\tabcolsep}{8pt}
    \begin{tabular}{ccc|ccc}
    \hline
    variant &biased prior &attention mechanism    & PSNR$\uparrow$ & SSIM$\uparrow$ & FID$\downarrow$ \\\hline
    MMT-1 &\Checkmark &concat      &  26.17    & 0.89  &  20.01  \\
    MMT-2 &\Checkmark &AdaIN~\cite{DBLP:conf/iccv/HuangB17}     &  26.17   & 0.89   &  21.71  \\
    MMT-3 &\Checkmark &SPADE~\cite{DBLP:conf/cvpr/Park0WZ19}     &  26.29   & 0.90  &   14.60  \\
    MMT-4 &\Checkmark &MSSA+ADN &  26.24    & 0.91     & 12.59   \\
    \hline
    MMT-5 &\XSolidBrush &MSSA+add  &  26.37    & 0.91     & 12.64   \\
    MMT-6 &\XSolidBrush &MSSA+conv &  26.50    & 0.91     & 11.90   \\
    MMT-7 &\XSolidBrush &MSSA+AdaIN~\cite{DBLP:conf/iccv/HuangB17} &  26.36    & 0.91   & 12.81   \\
    MMT-8 &\XSolidBrush &MSSA+SPADE~\cite{DBLP:conf/cvpr/Park0WZ19} &  26.42    & 0.91  & 12.17   \\
    MMT-9 &\XSolidBrush &MSSA+ADN &  \bf{26.70}    &   \bf{0.92}   &  \bf{11.40}   \\
    \hline
    \end{tabular}
\end{table}

{\flushleft {\bf Biased prior guidance.}}
Different from previous works~\cite{DBLP:conf/cvpr/XiongYLYLBL19,DBLP:conf/bmvc/SongYSWHK18,DBLP:conf/iccvw/NazeriNJQE19,DBLP:conf/eccv/LiaoXWLS20,DBLP:conf/cvpr/Liao00L021} relying on biased prior guidance from predicted auxiliary structures, we jointly learn the interplay information of multi-modal features across the three branches and guide image inpainting based on ADN. 
To demonstrate its effectiveness, we construct four variants that are directly guided by predicted auxiliary structures. In practice, we first add one convolutional layer at different stages to predict the auxiliary structures (Fig. \ref{fig:framework}), and then combine multi-modal features (Fig. \ref{fig:msa}).

In Table \ref{tab:abla_att}, MMT-1 denotes concatenating predicted structures with feature maps in the inpainting branch. MMT-2, MMT-3 and MMT-4 denote that we use AdaIN~\cite{DBLP:conf/iccv/HuangB17}, SPADE~\cite{DBLP:conf/cvpr/Park0WZ19}, and MSSA with ADN to calculate the affine transformation parameters $(\gamma,\beta)$ based on predicted structures, respectively. 
Compared with our method without biased prior guidance (\ie, MMT-9), the FID score is significantly reduced based on predicted auxiliary structures. The results support our statement that predicted structures may introduce additional noises in image inpainting intermediately without ground-truth.

{\flushleft {\bf Effectiveness of multi-scale spatial-aware attention.}}
To verify the effectiveness of Multi-Scale Spatial-aware Attention (MSSA), we construct four baseline feature fusion strategies from MMT-5 to MMT-8 in Table \ref{tab:abla_att}. MMT-5 means that we directly perform element-wise summation on features from three branches, while MMT-6 means that we splice the features from three branches together and then fuse them by two convolutional layers. 

From Table \ref{tab:abla_att}, our MSSA performs the best in terms of three metrics. Compared with simple addition or convolution, our MSSA can provide reliable cross-attention among multiple modalities to guide high-quality reconstructed images.
We also replace ADN by AdaIN~\cite{DBLP:conf/iccv/HuangB17} and SPADE~\cite{DBLP:conf/cvpr/Park0WZ19} in MSSA for MMT-7 and MMT-8 respectively. The results show that our ADN performs better than previous normalization methods, demonstrating its effectiveness.

{\flushleft {\bf Effectiveness of adaptive contextual bottlenecks.}}
In Table \ref{tab:abla_aed}, we compare our Adaptive Contextual Bottlenecks (ACB) with the vanilla ResNet block~\cite{DBLP:conf/cvpr/HeZRS16} and the recently proposed AOT~\cite{DBLP:journals/corr/abs-2104-01431}. ACB@$L$ ($L=2,4,6,8$) denotes $L$ layers of ACB modules; RES@8 and AOT@8 denote $8$ ResNet blocks~\cite{DBLP:conf/cvpr/HeZRS16} or $8$ AOT blocks~\cite{DBLP:journals/corr/abs-2104-01431} respectively. $\dag$ means quadrupling the channels of feature maps in ResNet blocks or copying base feature maps for different pathways in AOT blocks.
The results show that the performance of ACB is improved along with the number of blocks is increased from $2$ to $8$. Using $8$ ResNet or AOT blocks achieves similarly as that using $4$ ACB blocks. It is worth mentioning that ResNet and AOT blocks have less number of channels of feature maps in each pathway. For a fair comparison, we construct two variants $\dag$RES@8 and $\dag$AOT@8 with the same channels as our ACB blocks. However, more channels in feature maps do not help improve the performance by using ResNet or AOT blocks. We speculate that the gating updating scheme in our ACB can reduce the influence of redundant noisy context with more channels of feature maps.
 
Besides, the mean of category-wise intersection-over-union (mIoU)~\cite{DBLP:conf/eccv/ChenZPSA18} is another metric to validate the influence of bottleneck modules on segmentation inpainting. Our ACB module ($L\geq4$) still outperforms other two blocks by more than $2\%$. The segmentation results in Fig. \ref{fig:vis_abla_aed} also show that our ACB module generates more accurate segmentation performance. If the number of bottlenecks are increased, some isolated errors in segmentation can be removed (see the 3rd and 4th columns in Fig. \ref{fig:vis_abla_aed}).

\begin{table}[t]
\begin{minipage}[t]{0.48\textwidth}
    \makeatletter\def\@captype{table}
\scriptsize
\caption{Comparison between different bottlenecks.}
\label{tab:abla_aed}
\setlength{\tabcolsep}{1mm}{
\begin{tabular}{ccccc}
\hline
bottleneck & PSNR$\uparrow$ & SSIM$\uparrow$ & FID$\downarrow$ & mIoU\%$\uparrow$\\ \hline
RES@8  &  26.48    &  0.91    &  12.54 & 61.93 \\
$\dag$RES@8  &  26.23    &  0.91    &  13.26 &60.11  \\
AOT@8     &   26.51  &  0.91    &  11.61 & 63.68 \\
$\dag$AOT@8    &   26.29  &  0.91    &  14.17 &62.28  \\\hline
ACB@2     &   26.48   &  0.91    &  12.18 & 63.54 \\
ACB@4     &   26.60   &  0.91    &  12.24  & 65.84 \\
ACB@6     &   26.61   &  0.91    &   12.09 & 66.16 \\
\textbf{ACB@8}     &  \textbf{26.70}    &   \textbf{0.92}   &  \textbf{11.40} & \textbf{67.13} \\
\hline
\end{tabular}}
    \end{minipage}
\quad 
\begin{minipage}[t]{0.48\textwidth}
    \makeatletter\def\@captype{table}
    \caption{Efficiency of image inpainting networks.}
    \label{tab:model_speed}
    \tiny
    \centering
    \resizebox{\columnwidth}{14.5mm}{
    \begin{tabular}{ccccc}
    \hline
    method   &params (M) &MACs (G)    &speed (FPS) \\ \hline
    SPG~\cite{DBLP:conf/bmvc/SongYSWHK18}  & 119.64 & 58.68 & 2.03 \\
    EC~\cite{DBLP:conf/iccvw/NazeriNJQE19} & 27.06 & 122.67 & \bf{67.21} \\ 
    CTSDG~\cite{guo2021image} & 52.15 & \bf{17.67} & 36.99 \\
    RFR~\cite{DBLP:conf/cvpr/LiWZDT20} & 31.22 & 206.12 & 15.56 \\
    CSA~\cite{DBLP:conf/iccv/LiuJX019} & 132.11 & 55.23 & 1.37 \\ \hline
    RES@8~\cite{DBLP:conf/cvpr/HeZRS16} & \bf{22.76} & 96.10 & 40.82 \\
    AOT@8~\cite{DBLP:journals/corr/abs-2104-01431} & 27.48 & 100.93 & 30.96 \\ 
    Ours (ACB@2) & \bf{22.76} & 96.10 & 40.88                  \\
    Ours (ACB@8) & 51.09 & 125.11 & 29.49 \\ \hline
    \end{tabular}}
    \end{minipage}
\end{table}

{\flushleft {\bf Efficiency comparison.}}
From Table \ref{tab:model_speed}, we compare the number of parameters, computational complexity (MACs), and the running speed (FPS) of existing methods. Two-stage based SPG~\cite{DBLP:conf/bmvc/SongYSWHK18} and CSA~\cite{DBLP:conf/iccv/LiuJX019}, composed of complex sub-networks at each stage, run much more slowly than end-to-end methods. In contrast, EC~\cite{DBLP:conf/iccvw/NazeriNJQE19} consists of two simple sub-networks for edge prediction and image inpainting, resulting in fast running speed but inferior performance. RFR~\cite{DBLP:conf/cvpr/LiWZDT20} is an end-to-end model but predicts the inpainted results by the decoding heads recurrently.
In terms of bottlenecks in the encoder, our ACB@2 achieves similar performance as AOT@8 with faster speed. By using $8$ blocks, our method is still efficient with state-of-the-art performance among end-to-end methods. 

{\flushleft {\bf Limitation discussion.}}
Although our model generates promising results in most cases, it fails to recognize and recover unseen semantic knowledge, hence produces strange artifacts in complex scenes with large masks. Note that this weakness also affects other methods. It indicates that image inpainting model requires not only generative but also recognition capability. For example, our method can synthesize the human silhouette but lacks precise semantic details.

\section{Conclusion}
In this paper, we propose an end-to-end Multi-modality Guided Transformer for image impainting, which enriches coupled spatial features from shared multi-modal representations (\ie, RGB image, semantic segmentation and edge textures). The proposed Multi-Scale Spatial-aware Attention can integrate compact discriminative features from multiple modalities via Auxiliary DeNormalization. Meanwhile, we introduce the Adaptive Contextual Bottlenecks in the encoder to enhance context reasoning for more semantically consistent inpainted results for the missing region. To the best of our knowledge, our scientific value lies in first analyzing the biased prior problem in image inpainting.

{\flushleft {\bf Acknowledgements and Declaration of Conflicting Interests.}} This work was supported by the Key Research Program of Frontier Sciences, CAS, Grant No. ZDBS-LY-JSC038. Libo Zhang was supported Youth Innovation Promotion Association, CAS (2020111). Dr. Du and his employer received no financial support for the research, authorship, and/or publication of this article.

%
%
\bibliographystyle{splncs04}
\bibliography{reference}
\end{document}